Taylor & Francis
Taylor & Francis Group

# On the use of adversarial validation for quantifying dissimilarity in geospatial machine learning prediction

Yanwen Wang , Mahdi Khodadadzadeh and Raúl Zurita-Milla

Faculty of Geo-Information Science and Earth Observation (ITC), University of Twente, Enschede, The Netherlands

## ABSTRACT

Recent geospatial machine learning studies have shown that the results of model evaluation via cross-validation (CV) are strongly affected by the dissimilarity between the sample data and the prediction locations. In this paper, we propose a method to quantify such a dissimilarity in the interval 0 to 100% and from the perspective of the data feature space. The proposed method is based on adversarial validation, which is an approach that can check whether sample data and prediction locations can be separated with a binary classifier. The proposed method is called dissimilarity quantification by adversarial validation (DAV). To study the effectiveness and generality of DAV, we tested it on a series of experiments based on both synthetic and real datasets and with gradually increasing dissimilarities. Results show that DAV effectively quantified dissimilarity across the entire range of values. Next to this, we studied how dissimilarity affects CV methods' evaluations by comparing the results of random CV method (RDM-CV) and of two geospatial CV methods, namely, block and spatial+ CV (BLK-CV and SP-CV). Our results showed the evaluations follow similar patterns in all datasets and predictions: when dissimilarity is low (usually lower than 30%), RDM-CV provides the most accurate evaluation results. As dissimilarity increases, geospatial CV methods, especially SP-CV, become more and more accurate and even outperform RDM-CV. When dissimilarity is high ($\geq 90\%$), no CV method provides accurate evaluations. These results show the importance of considering feature space dissimilarity when working with geospatial machine learning predictions and can help researchers and practitioners to select more suitable CV methods for evaluating their predictions.

## 1. Introduction

Machine learning (ML) is widely used in geospatial prediction to estimate unknown values at specific prediction locations (Aguilar et al. 2018; Hengl et al. 2018; Usman et al. 2023). These predictions are often done to create spatially continuous products, for example, mineral (Khodadadzadeh and Gloaguen 2019), health risk (Garcia-Marti et al. 2018), or phenological (Zurita-Milla, Laurent, and van Gijsel 2015) maps. In these and many other applications, predictions come from ML regression models trained on limited sample data, where the number of samples is typically much smaller than the number of prediction locations.

This imbalance between samples and prediction locations is mostly due to practical limitations such as accessibility (Lamichhane, Kumar, and Wilson 2019) or sampling costs (Hengl et al. 2015). For similar reasons, collecting additional data for an independent evaluation of geospatial ML prediction is rarely feasible (Valavi et al. 2019). To address these operational constraints, the evaluation of geospatial ML models is mainly conducted by splitting the available sample data into training and validation subsets (de Bruin et al. 2022; Y. Wang, Khodadadzadeh, and Zurita-Milla 2023). Random k-fold cross-validation (RDM-CV) stands out as the most popular evaluation method (G. Chen et al. 2018; Guo et al. 2022; Nesha et al. 2020). As the name indicates, RDM-CV randomly splits the sample data into k equal-size folds, and then, it iteratively uses one of them as a validation subset and the remaining ones as a training subset. When sample data are randomly or regularly collected over the entire study area (Brus, Kempen, and Heuvelink 2011; Lagacherie et al. 2020; J. F. Wang et al. 2012), RDM-CV can provide sufficiently accurate evaluation results (Milà et al. 2022; Wadoux et al. 2021). This is because, under these circumstances, the training and validation subsets are representative of the relationship between sample data and

---

**CONTACT** Yanwen Wang  y.wang-4@utwente.nl

prediction locations. Specifically, random sampling and regular sampling ensure that the sample data and prediction locations are similar from the perspective of data distribution, whilst the random split of RDM-CV can also guarantee that the training and validation subsets are similar.

In practical situations, most geospatial ML predictions can only be collected from limited regions of the entire study area, potentially leading to significant differences between the sample data and the prediction locations. A representative case is the large-scale prediction (S. Chen et al. 2022; Ludwig et al. 2023; Mussumeci and Codeço Coelho 2020) where sample data are often concentrated on a few developed and accessible regions (Meyer and Pebesma 2022); for instance, global soil maps are produced with sample data clustering among Europe and North America (Guerra et al. 2020). Another case is making predictions in a completely new area. For example, the predictions of the affected area after an earthquake are so urgent that collecting samples is almost unfeasible (B. Li et al. 2021); other examples are the predictions of landslides (Goetz et al. 2015; Y. Li et al. 2021; Zhao et al. 2017) or the predictions of invasive species diffusion (Cheng et al. 2018), where collecting samples in the study area is also impossible, as the target phenomena have not occurred yet. In all the above cases, geospatial ML acts as an extrapolation model for predicting values that extend beyond the known data (i.e. training data).

In the scenarios discussed above where the sample data and prediction location are different, RDM-CV tends to be over-optimistic and not suitable for evaluation (Brenning 2005; Pohjankukka et al. 2017; Stock and Subramaniam 2022; Wiens et al. 2008). Consequently, a series of geospatial CV methods have been proposed with the core idea of avoiding excessive similarity between the training and validation subsets. Block CV (BLK-CV) and spatial+ CV (i.e. spatial-plus CV and SP-CV) are two representative methods in this regard. BLK-CV has a long history (Brenning 2012; Roberts et al. 2017; Valavi et al. 2019) and is widely used in evaluation (Bueno, Macera, and Montoya 2023; Wadoux et al. 2021; Y. Wang, Khodadadzadeh, and Zurita-Milla 2023). As its name implies, BLK-CV would divide the sample data into contiguous blocks and then randomly split blocks (instead of samples) as k-folds. SP-CV is a recently proposed geospatial CV method

(Y. Wang, Khodadadzadeh, and Zurita-Milla 2023) that considers the feature space. In SP-CV, agglomerative hierarchical clustering (AHC) is used first to divide samples into improved blocks. Then, all blocks are split into folds by cluster ensembles based on their locations, covariates, and the target variable. As shown in Y. Wang, Khodadadzadeh, and Zurita-Milla (2023), SP-CV shows promising evaluation results when sample data and prediction locations are substantially different.

According to the above descriptions of RDM-CV and geospatial CV methods, it can be observed that the dissimilarity (or similarity) between the sample data and the prediction locations is a decisive factor for determining the evaluation accuracy of CV methods (for brevity, dissimilarity between samples and prediction locations will be abbreviated as dissimilarity in most cases). This has been confirmed by recent studies (de Bruin et al. 2022; Milà et al. 2022; Wadoux et al. 2021). It should be noticed that the transition from largely similar to substantially different is gradual. For example, varying degrees of samples clustering in the prediction area would result in different degrees of dissimilarity (Milà et al. 2022). Therefore, here we use dissimilarity as a continuous attribute to describe the relationship between sample data and prediction locations. Although a few studies recognized this and considered dissimilarity when proposing new CV methods (e.g. Meyer and Pebesma (2022) and Linnenbrink et al. (2024)), they have not explicitly expressed and quantified the dissimilarity between the sample data and the prediction locations.

In this paper, we propose a novel method that introduces adversarial validation (AV) to quantify the dissimilarity between samples and prediction locations for geospatial ML predictions, which we name dissimilarity quantification by adversarial validation (DAV). Additionally, another key contribution of this paper is the experimental comparison of CV methods based on DAV. Through numerous experiments with gradually changing dissimilarity scenarios, we investigate the relationship between dissimilarity and the evaluations of random CV and geospatial CV methods in detail. The experimental results presented in this paper provide important insights that complement previous studies, which have considered only a few dissimilarity scenarios.

The remainder of this paper is organized as follows: In Section 2, we specify the proposed method to quantify the dissimilarity and introduce CV methods compared in the experiments. In Section 3, we describe and discuss our experiments and results. Finally, in Section 4, we present the main conclusions of this study and provide recommendations for future research.

## 2. Methods

In this research, we aim at proposing a method to quantify the dissimilarity. In Subsection 2.1, we introduce this proposed method (DAV) in detail. Our research also includes investigating the impact of dissimilarity on CV methods. Therefore, we introduce the CV methods used in experiments in Subsection 2.2.

### 2.1. Dissimilarity quantification by adversarial validation

AV is a technique proposed by FastML (2016) to detect and mitigate the problems of data distribution differences between test and training datasets. The core idea of AV is treating test and training datasets as separate categories in a binary classification problem (FastML 2016; Zhang et al. 2023). As Figure 1 shows, test and training sets are labeled as 0 and 1, respectively, and then a classifier is trained to distinguish between them. If the classifier struggles to differentiate between the test and training data (indicated by a low classification accuracy), it suggests that the two sets are from the same data distribution. Conversely, if the classifier can easily separate the two sets (reflected by a high classification accuracy), it indicates that the training and test sets have different distributions. Next, depending on the accuracy of the classifier, specific steps can be taken to solve the problems caused by data distribution differences, for example, selecting the training data most similar to the test data as the validation subset to get a more accurate error estimation (Ishihara, Goda, and Arai 2021) and removing the top contributing features of the classifier to improve the generalization ability of the ML prediction model (Pan et al. 2020).

In this research, we adopt the AV technique to quantify the dissimilarity between samples and prediction locations. It is better than other possible ways (such as directly calculating the Euclidean distance in the feature space) because the classifier of AV is able to capture complex and nonlinear relationships between two datasets. To the best of our knowledge, this is the first time that AV has been applied to geoscience, especially to the domain of evaluating geospatial ML prediction. In addition, we have addressed the following special issues in our proposed DAV. First, samples and prediction locations commonly have the number imbalance problem. Second, we should compute a percentage value to quantitatively represent the degree of dissimilarity. Third, our purpose of quantifying dissimilarity based on AV is to investigate the impact of dissimilarity on the evaluation performances of CV methods, rather than selecting validation subsets or removing features (which are common in previous studies). Figure 2 shows the basic workflow of DAV and Algorithm 3 provides the pesudo-code of DAV to present the specific steps.

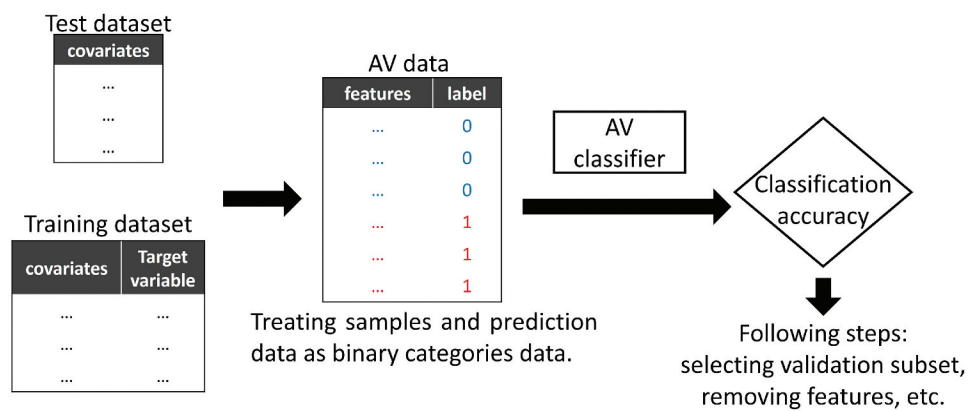

**Figure 1.** Adversarial validation (AV) schematic diagram.

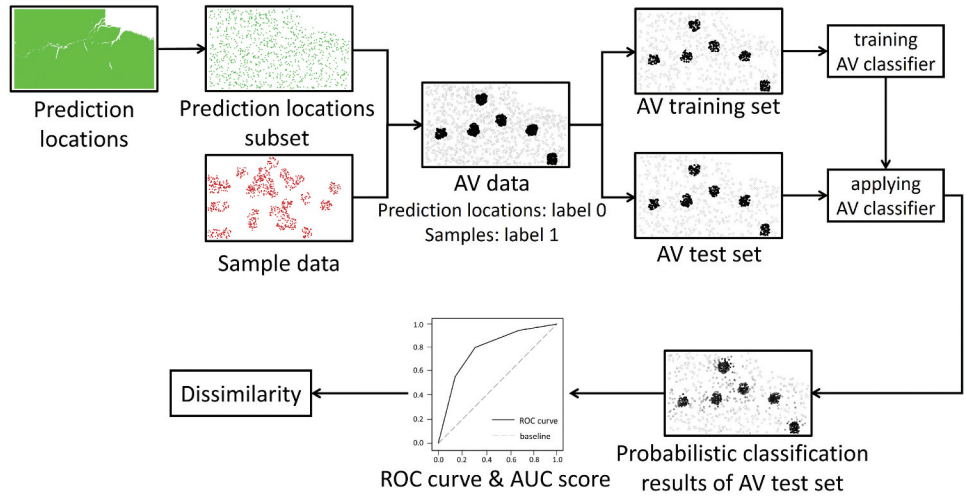

**Figure 2.** The workflow of dissimilarity quantification by adversarial validation (DAV).

---

**Algorithm 1 Dissimilarity quantification by adversarial validation (DAV)**

**Input**: A samples set, $Data_N^{sample}$ (where $N$ is the number of samples locations); A prediction set, $Data_M^{pred}$ (where $M$ is the number of prediction locations); an adversarial classifier, **AVclassifier**.

1: **Step 1**: $Data_N^{pred} = \mathbf{Rand}(N, Data_M^{pred})$ (∗ **Rand** randomly selects a number of locations equal to the number of samples ∗)

2: **Step 2**: Construct AV data

3: $Data_N^{pred}.\mathbf{add}(\mathbf{0})$ (∗ add a label 0 column to the prediction set ∗)

4: $Data_N^{sample}.\mathbf{pop}(\mathbf{target})$ (∗ remove the original target variable from the samples set ∗)

5: $Data_N^{sample}.\mathbf{add}(\mathbf{1})$ (∗ add a label 1 column to the samples set ∗)

6: $Data_{2N}^{AV} = Data_N^{pred} \cup Data_N^{sample}$ (∗ combine the prediction locations and samples sets and construct the AV dataset ∗)

7: **Step 3**: Split AV dataset into a training and test subsets.

8: **Shuffle**$(Data_{2N}^{AV})$ (∗ randomly shuffle the AV dataset ∗)

9: $Data_N^{AV_{train}} \leftarrow Data_{2N}^{AV}[1 : N]$

10: $Data_N^{AV_{test}} \leftarrow Data_{2N}^{AV}[N+1 : 2N]$

11: **Step 4**: Train the adversarial classifier

12: **AVclassifier**.train$(Data_N^{AV_{train}})$

13: **Step 5**: Apply the adversarial classifier to classify the AV test subset

14: $\hat{P}^{AVtest} \leftarrow$ **AVclassifier**.predict$(Data_N^{AV_{test}})$ (∗ obtain classification probabilities ∗)

15: **Step 6**: Draw the ROC curve and calculate the AUC score

16: $AUC_{score} \leftarrow$ ROC $-$ AUC$(Data_N^{AV_{test}}, \hat{P}^{AVtest})$

17: **Step 7**: Obtain Dissimilarity value

18: $D \leftarrow Normalize(AUC_{score})$ (∗ normalize according to Equation 1 ∗)

**Output**: Dissimilarity, $D$.

---

In geospatial ML prediction, the number of prediction locations is typically much larger than the number of samples (Meyer and Pebesma 2021). This can lead to the problem of class imbalance, making it difficult for the AV classifier to notice the sample data. Therefore, we need to select a subset of prediction locations as the same number of samples to avoid class imbalance, and this subset of prediction locations should represent all prediction locations unbiasedly. We employ a random selection in step 1 to conduct such an unbiased representation (Brus, Kempen, and Heuvelink 2011; Wadoux et al. 2021).

Step 2 is constructing AV data, which requires to transform the samples and the prediction locations' subset into binary categories' data. Specifically, we keep the covariates of samples and prediction locations' subset unchanged and then label prediction locations as 0 and samples as 1 following the study of Qian et al. (2022). They are combined together to form the AV data. Figure 3a is an example of AV data. In this example, samples are strongly clustered, i.e. black dots are concentrated groups in Figure 3a.

Next, the AV data should be split into AV training and test sets in step 3. Both AV training and test sets need to unbiasedly represent AV data. Such unbiased representation is necessary for guaranteeing that both AV training and test sets can well reflect the samples and prediction locations together, and then ensuring that the classifier trained in step 4 can adequately address the data of two categories (samples and prediction locations) simultaneously. Therefore, we randomly split AV data to obtain training and test sets. The examples of AV training set and AV test set are shown by Figures 3b,c respectively. According to the example shown in Figure 3, we can see that both AV training and test sets have enough and almost equal-sized prediction location category data (label 0) and sample category (label 1) data, and the spatial distributions of their data are both highly similar with complete AV data. It indicates that both AV training and AV test sets achieve good representations.

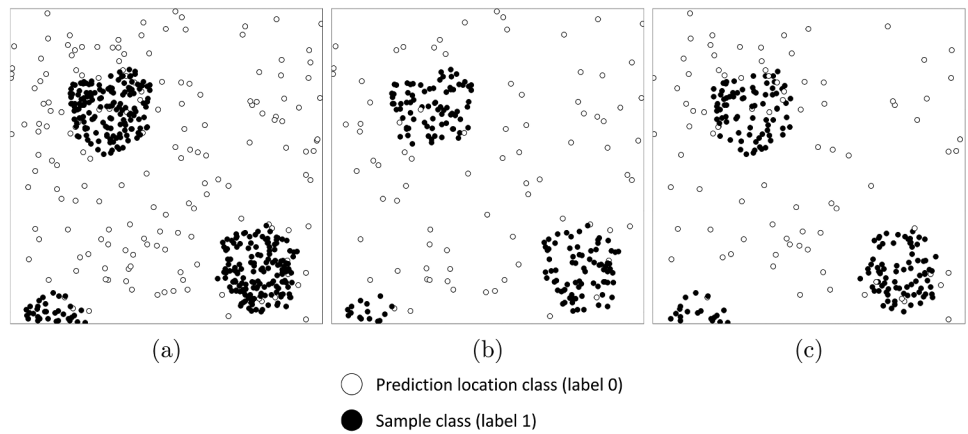

**Figure 3.** An example of AV data. (a). The complete AV data. (b). AV training set. (c). AV test set.

In step 4, we use AV training data to train an AV classifier, which is the key component of AV and DAV. The AV classifier is a binary ML classifier to discriminate two datasets (Pan et al. 2020). If the classification by the AV classifier is ideal, i.e. two datasets can be easily distinguished, it indicates that they are largely dissimilar. Conversely, a failure of the AV classifier means that they are hardly distinguished, indicating that they are quite similar. In DAV, the AV classifier is used to quantify the dissimilarity between prediction locations and samples. Hence, the target variables used in the AV classifier of DAV are the binary categories constructed in step 2 (with samples labeled as 1 and prediction locations labeled as 0). For the input features, since DAV is used to quantify the dissimilarity in a specified ML prediction task, we opt for the same with the ones of prediction task. Specifically, we feed all input features of the ML prediction model into the AV classifier with the same elements, value ranges, order, and all contents.

The AV classifier is trained by the constructed AV training set in step 3 by an ML model, and then it is applied to the AV test set (also from step 3) to calculate the dissimilarity in the following steps. Note that in theory, any ML algorithm that can be used as a binary classifier is acceptable. However, in order to ensure the rationality of the dissimilarity quantified by DAV, it is necessary to carefully choose the classifier. Furthermore, it is preferable that the ML algorithm (including hyperparameters) for the AV classifier and for the prediction task are the same. This helps to guarantee that the quantified dissimilarity can better match the corresponding geospatial ML prediction task.

In this research, random forest (RF) is used both as the AV classifier and as the ML model for the geospatial prediction (regression) task. The following reasons justify our choices. First, choosing RF helps to keep a consistent and comparable work with related studies. RF has consistently been used as the only prediction model in previous research of dissimilarity and CV methods (de Bruin et al. 2022; Milà et al. 2022; Wadoux et al. 2021). Second, choosing RF can ensure that DAV has good stability. Since dissimilarities widely exist in various geospatial ML predictions and DAV has to quantify such diverse dissimilarities, the AV classifier of DAV should have stable performances across different datasets. Simple ML models are usually better than complex ML models in terms of stability and can avoid overfitting problems to a certain extent. Hence, a simple model is typically used as the AV classifier (Montesinos-López, Montesinos-López, and Montesinos-López 2023; Qian et al. 2022), just like using RF as the AV classifier in Pan et al. (2020). In addition, RF has shown good performance and stability in dealing with noise and outliers (Liaw and Wiener 2002). RF is user-friendly (Hengl et al. 2018) and efficient (Habibi et al. 2023), especially it can be parallelized to further improve the computation-efficiency (Guan et al. 2013). Hence, choosing RF as the AV classifier and the prediction model also facilitates the implementation of this research.

In this research, we used the Python library scikit-learn (version 1.0.2) to build the RF model. We set two key hyperparameters of RF: *Ntree* (the number of decision trees) to 500 and *Mtry* (the number of covariates chosen for the best split) to the square root of the number of covariates. As for the other hyperpara-meters, we set *max_depth* to "None," meaning the all nodes are expanded until all leaves are pure (i.e. samples within the same node have the same label, or node has only a single sample), *min_samples_split* to two, *max_leaf_nodes* to unlimitation, and *bootstrap* to TRUE.

After training the AV classifier, we use it to classify every data point of the AV test set in step 5. This is for the following calculation of classification accuracy in step 6, and the classification accuracy is the basis of quantified dissimilarity. By averaging the classifica-tions of all decision trees of RF, the AV classifier can get the probabilistic results for all AV test data (Belgiu and Drăguţ 2016). The reason for using probabilistic classification is that AV represents classifier accuracy with the AUC score, and the calculation of the AUC score requires probabilistic classification results. The calculation of the AUC score will be introduced in step 6. In addition, using probabilistic classification is superior to hard classification (i.e. directly using the predicted labels of AV test data), because probabilistic classification is unaffected by threshold settings and can provide uncertainty information. This helps con-tribute to the rationality and robustness of the next calculated classification accuracy. Therefore, probabil-istic classification is adopted by AV (Montesinos-López, Montesinos-López, and Montesinos-López 2023), and we also apply it in DAV.

Based on the classification results of the AV test set, we can calculate the accuracy of the AV classifier in step 6. In AV, the Area Under Curve (AUC) score is calculated, i.e. the area under the Receiver Operating Characteristic (ROC) curve, to depict the accuracy. The ROC curve plots the true positive rate against the false-positive rate of classification results at various threshold settings. Therefore, it requires the probabil-istic classifications. Then, the AUC score quantifies the accuracy of the classifier by measuring the area under the ROC curve. The AUC score considers both the true positive rate (TPR) and the false-positive rate (FPR) at different classification thresholds. It means that the variation of classification thresholds does not affect the calculation of the AUC score, thus enabling a more reliable reflection of the classifier accuracy. The AUC score is also widely used in geospatial ML predictions (J. Chen et al. 2024; Hitouri et al. 2022). A higher AUC value indicates that the AV classifier is more accurate (Wu et al. 2019) and also implies that the dissimilarity between samples and prediction locations is larger. The value range of AUC is usually [0.5, 1], but some-times, the AUC might be slightly lower than 0.5. An AUC value of 0.5 means that the classifier is almost randomly guessing if data belongs to class 0 or 1, indicating that sample data and prediction locations have the same data distributions.

Because using 0.5 as the minimum value of dissim-ilarity is confusing, in step 7, we normalize the AUC score to a new metric, directly named as dissimilarity ($D$). The normalization function is shown by Equation 1. $D$ is a percentage value within the range of [0, 100%].

$$D = \begin{cases} (\frac{AUC\,score - 0.5}{1 - 0.5}) \times 100\%, & if\ AUC\ score > 0.5 \\ 0\%, & if\ AUC\ score \leq 0.5 \end{cases}$$

(1)

## 2.2. Cross-validation methods

In this section, we introduce three CV methods (which are used in the experiments presented in this paper): RDM-CV, BLK-CV, and SP-CV. RDM-CV and BLK-CV are the most commonly studied and compared CV meth-ods in the research on evaluating geospatial ML pre-dictions (Milà et al. 2022; Roberts et al. 2017; Wadoux et al. 2021). SP-CV is our recently proposed geospatial CV method (Y. Wang, Khodadadzadeh, and Zurita-Milla 2023). SP-CV split samples by considering both the geographic and feature spaces. In our previous work, we showed that SP-CV can produce more rational fold splits and more accurate evaluation results compared to the commonly used geospatial CV methods.

The distinctions of RDM-CV, BLK-CV, and SP-CV lie in their fold splits. Figure 4 shows the examples with 100 samples of three CV methods' 5 folds, which can help us to introduce and compare their folds splits. RDM-CV randomly splits samples into $k$ equal-size folds. We can see that samples of each fold are all randomly and evenly distributed across the entire study area in Figure 4a. In BLK-CV, the study area should be divided into contiguous square blocks at first. The side length of the block is typically set as the spatial autocorrelation threshold (Roberts et al. 2017), which can be calculated by the semi-variogram of samples' target variable values. After that, the divided blocks instead of individual samples are randomly split into $k$ folds. As Figure 4b shows, the samples within the same block are forced to the same fold, which can help avoid spatial autocorrelation in the evaluation to a certain extent (Ploton et al. 2020). The process of splitting folds of SP-CV is more complex and has two steps. The first step is using agglomerative hierarchical clustering (AHC) to divide samples into blocks. Compared with BLK-CV blocks, the blocks of SP-CV have considered the spatial distribution of samples. Thus, as Figure 4c shows, the samples of SP-CV blocks are in the center of each block. Next, the second step is using the clusters ensemble (CE) to split blocks into $k$ folds, where CE is based on the clusters of samples' coordinates, covariates, and target variables, respectively. Therefore, the folds of SP-CV can better reflect the dissimilarity of data feature space (Y. Wang, Khodadadzadeh, and Zurita-Milla 2023). That is why we can see that folds of SP-CV are much less randomized than RDM-CV and BLK-CV in Figure 4.

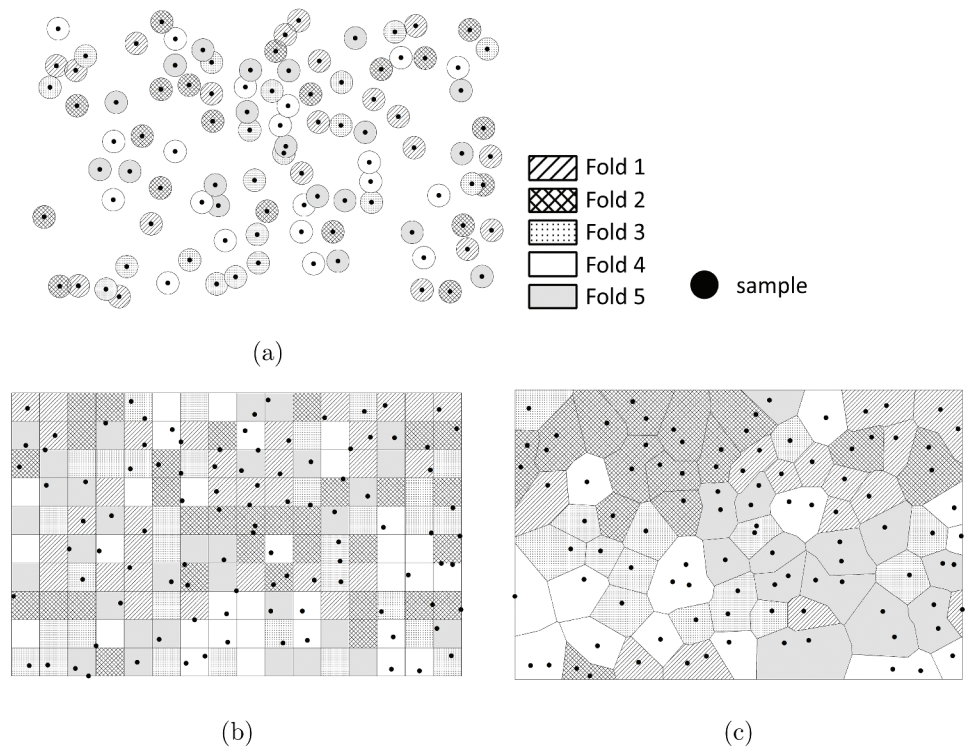

(a)

(b)                                          (c)

**Figure 4.** Examples of CV methods folds split (100 samples and 5 folds). (a) RDM-CV. (b) BLK-CV. (c) SP-CV.

The procedures for carrying out the evaluation by RDM-CV, BLK-CV, and SP-CV are basically the same. The first step is to split all samples into $k$ folds. In this research, $k$ is set to 5 because it is a commonly used number of folds (Lyons et al. 2018). The second step involves $k$ rounds of validation, with each fold serving as the validation subset iteratively and the remaining folds comprising the training subset. In each validation round, the specified ML algorithm is trained on the training subset to obtain a prediction model, and then, this model is used to predict all samples of the validation subset. After all validation rounds, every sample will have a predicted value of the target variable. The third step is to calculate the evaluation metric. Based on the true and predicted values (of the target variable) of all samples, a statistical metric can be calculated to describe the prediction error. This prediction error calculated by the CV method is an estimate of the actual error for the corresponding geospatial ML prediction, i.e. it is the evaluation of the geospatial ML prediction by this CV method. In this research, we use the root-mean-square error (RMSE) as the evaluation/accuracy metric of CV methods, because it is one of the most widely used statistical metric for describing prediction error (Oliveira, Torgo, and Santos Costa 2021), especially in geospatial CV methods' studies (Ploton et al. 2020; Roberts et al. 2017).

## 3. Experiments

To study the effectiveness of DAV and investigate the impact of dissimilarity on CV methods, we designed a series of experiments using synthetic and real datasets. The following subsections describe the datasets, our experimental setup, and results.

## 3.1. Datasets

We use two datasets for the experiment. The first dataset is a synthetic dataset, which is developed by Roberts et al. (2017) as an ecological prediction case. This dataset contains seven covariates and a target variable (i.e. species abundance). However, only three covariates are related to the synthetic target variable, while the remaining four covariates are completely unrelated to the synthesizing process of the target variable. This settlement is to simulate the situation that the prediction ability of the geospatial ML model is restricted. All these variables are generated over a 1000 × 1000 raster layer. Most covariates are created based on Gaussian Random Field (GRF) (Schlather et al. 2015) to simulate the actual spatial variables' autocorrelation structures (Le Rest et al. 2014; Sarafian et al. 2021). There is also a regional covariate generated by the Markov random field (MRF) to simulate regional patterns of geoscience covariates. The second dataset is a real dataset of above ground biomass (AGB) in the Brazil Amazon basin. It is adopted from the study of Wadoux et al. (2021). This dataset has 28 covariates, and the AGB target variable. Unlike the synthetic dataset, all covariates of the real dataset are related to AGB and carefully collected to maximize the prediction ability of the Amazon AGB prediction model. All the covariates and the target variable are available as raster layers with a resolution of 1 × 1 km. Figure 5 shows two datasets and their target variables' distributions. Detailed information on two datasets are included in Appendix 1.

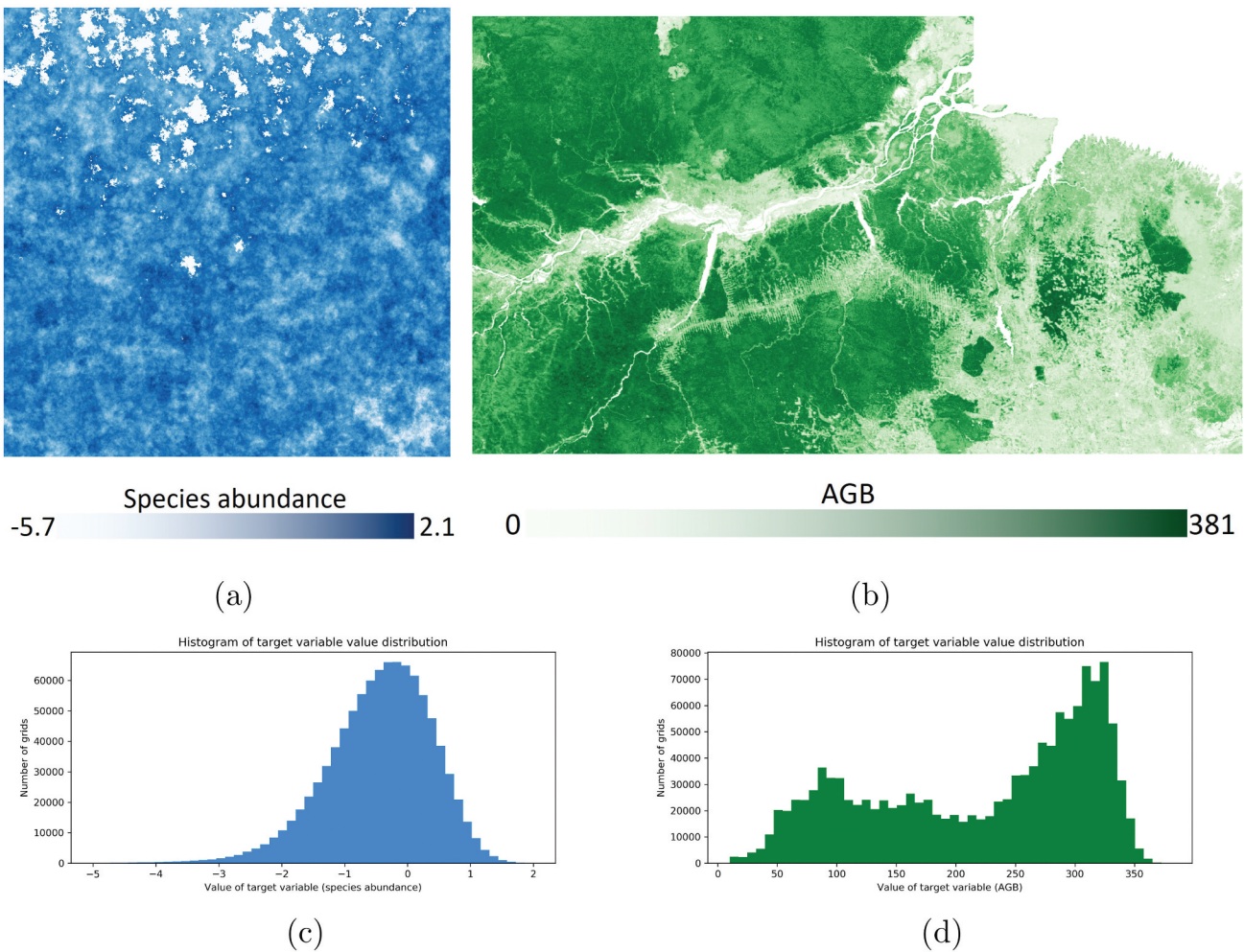

**Figure 5.** Datasets of the experiment. (a) Synthetic species abundance dataset. (b) Real amazon AGB dataset. (c) Data distribution of synthetic species abundance. (b) Data distribution of AGB.

The key factor to ensure the effectiveness of the experiment is whether it includes sufficient prediction tasks with diverse dissimilarity levels. Consequently, the selected dataset (especially the target variable) for implementing the experiment should exhibit a clear spatial heterogeneity structure. Only a spatially heterogeneous dataset can ensure that clustered samples (i.e. samples concentrated in limited regions of the study area) and prediction locations (i.e. the entire study area) have a certain degree of dissimilarity. Furthermore, only with a spatially heterogeneous dataset, we can construct the continuously changing dissimilarity prediction tasks by controlling the clustering level of the samples, enabling us to test the effectiveness of DAV across a range of diverse dissimilarity scenarios. As shown in Figure 5, both synthetic and real datasets have clear spatial heterogeneous structures. Figure 5a shows that the target variable (species abundance) of the synthetic data set commonly has lower values around the lakes (that is, the blank regions of the north part with no data in Figure 5a) and the south-east corner, while the value is clearly higher in the south-west corner. In comparison, Figure 5b shows that AGB of the real dataset has a more obvious spatial heterogeneity structure: AGB is substantially lower among the banks of the Amazon river and in the South West regions with abundant human activities (like farming and urbanization), while the AGB is naturally much higher in the remaining rainforest regions. Therefore, based on these two spatial heterogeneous datasets, we could construct the experiment with sufficient dissimilarity degrees to verify whether DAV is effective or not.

### 3.2. Experiments and results

As Figure 6 shows, the experiments in this research are composed of three steps. Step 1 constructs the geospatial ML prediction tasks with gradually increasing dissimilarities. In step 2, we calculate the dissimilarities and the corresponding CV methods' evaluation performances of all prediction tasks. Finally, by step 3, the results of all prediction tasks are put together as scatter plots to reveal the relationship between dissimilarity and CV methods' evaluations.

#### 3.2.1 Step 1: construct prediction tasks with gradually changing dissimilarities.

A single prediction task only corresponds to one dissimilarity and one evaluation result of each CV method. To investigate how dissimilarity affects the evaluation performances of CV methods, we need a large number of prediction tasks with gradually changing dissimilarities. Therefore, in this research, we need to artificially construct prediction tasks based on the aforementioned two datasets. For each dataset, since its study area and data are fixed, the different dissimilarities should be reflected through different samples. In other words, the essence of constructing a prediction task is to determine the samples' set. Here, we adopted a commonly used approach to generate a series of samples' locations (de Bruin et al. 2022; Milà et al. 2022; Wadoux et al. 2021), with each set of samples corresponding to a specific prediction task.

First, the number of samples in all prediction tasks is set to be constant. Following other studies (Amato et al. 2020; Sarailidis, Wagener, and Pianosi 2023), it is set to 1000. Then, as in Wadoux et al. (2021), the study area is divided into 100 subregions by K-means clustering based on raster grids' coordinates. Thirdly, a number of subregions are randomly selected. Finally, we equally and randomly select all samples only from the selected subregions. After all these steps, a specific sample set is determined, i.e. a specific prediction task is constructed. In terms of the dissimilarity of this prediction task, the more subregions selected, the larger the proportion of the study area covered by samples, and consequently the lower the dissimilarity between the samples and the prediction locations. Therefore, for each dataset, the selected subregions are continuously increased from 1 to 100 to ensure that the constructed prediction tasks have gradually changing dissimilarities. In addition, to reduce random errors, the sampling of each specified number of selected subregions is repeated 10 times. Therefore, the amount of all constructed predictions for a dataset is 100 × 10 = 1000. Figure 7 shows some examples of constructed predictions for the synthetic and real datasets.

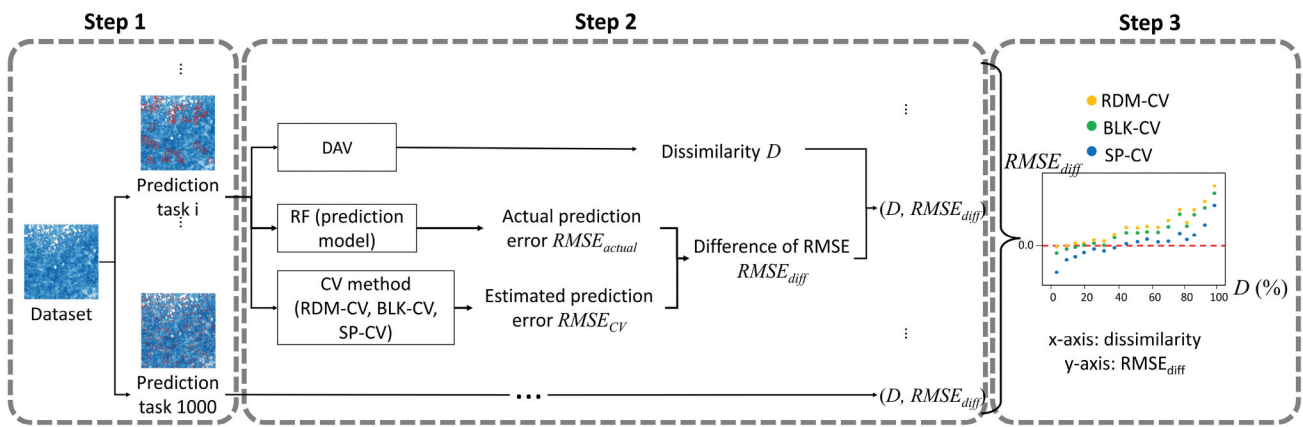

**Figure 6.** The workflow of the experiments for studying the relationship between dissimilarity and CV method evaluation performance.

### 3.2.2 Step 2: Calculate dissimilarities and CV methods' evaluation performances.

After constructing a series of prediction tasks, we need to calculate the dissimilarity of each prediction task and the corresponding evaluation performances of three CV methods (RDM-CV, BLK-CV, and SP-CV) to investigate their relationships. Therefore, in step 2 of experiments, we first use DAV to calculate the dissimilarities and then obtain the evaluation performances by conducting the CV methods on samples.

Figure 8 shows the quantified dissimilarities of the constructed prediction tasks, containing the plots of the number of selected subregions vs the dissimilarity value and the histograms of dissimilarity. To be specific, in our experiment, we controlled the clustering level to simulate the specified dissimilarity degree. In Figure 8 of both synthetic and real datasets' experiments, the quantified dissimilarity values are completely distributed among the entire range of [0, 100%]. In addition, the scatter plots clearly show that dissimilarity and samples-covered-area (i.e. the number of subregions for selecting samples) are negatively related, that is, the dissimilarity and clustering level are clearly positively related, consistent with previous research conclusions (Milà et al. 2022; Wadoux et al. 2021). Together, these results demonstrate that the quantified dissimilarities match the constructed prediction tasks, showing that DAV effectively quantified the dissimilarity in the experiments.

The evaluation performance of a CV method is essentially the difference between the actual prediction error of the ML prediction task ($RMSE_{actual}$) and

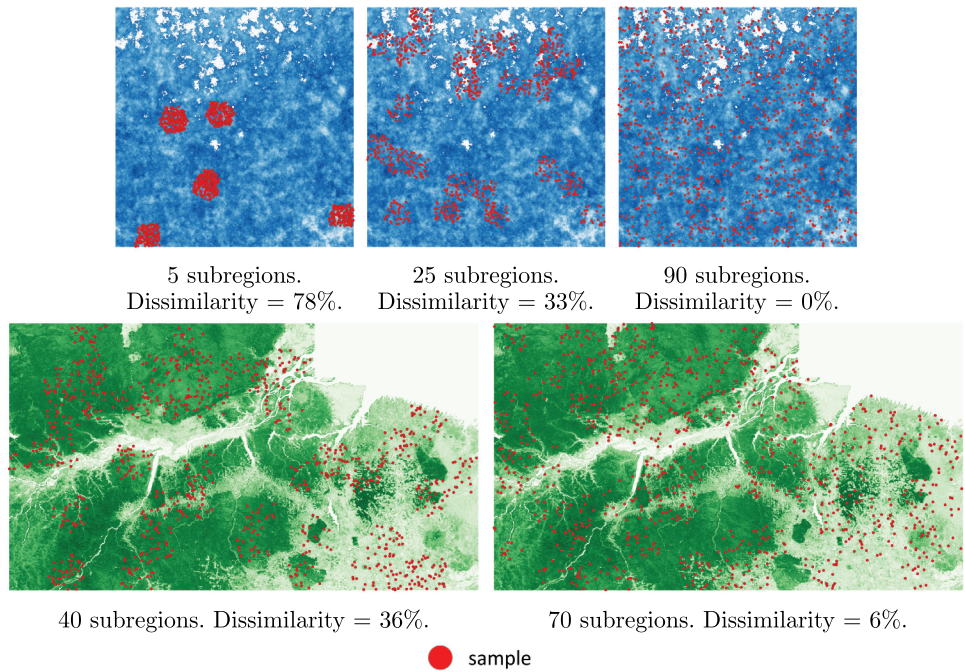

5 subregions.          25 subregions.          90 subregions.
Dissimilarity = 78%.   Dissimilarity = 33%.    Dissimilarity = 0%.

40 subregions. Dissimilarity = 36%.        70 subregions. Dissimilarity = 6%.

🔴 sample

**Figure 7.** Examples of constructed prediction tasks with $N$ selected subregions.

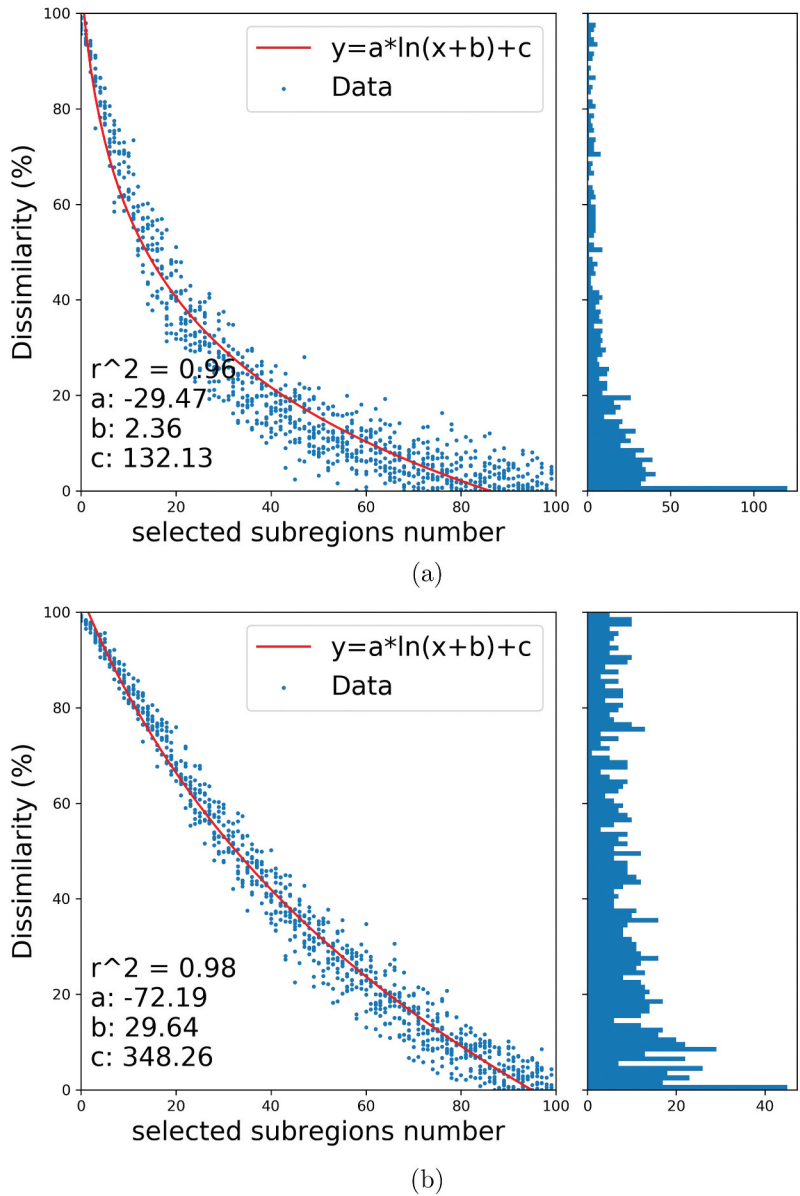

(a)

(b)

**Figure 8.** The dissimilarities of constructed prediction tasks. Left: scatter plots of number (*N*) of selected subregions (x-axis) vs dissimilarities (y-axis). Right: Histograms of frequency (x-axis) vs dissimilarities (y-axis). (a). Synthetic dataset. (b). Real dataset.

the estimated prediction error of that CV method ($RMSE_{CV}$) (Milà et al. 2022; Wadoux et al. 2021). For brevity, the prediction performance is usually abbreviated as $RMSE_{diff}$ (Milà et al. 2022). The calculation of $RMSE_{diff}$ is shown by Equation 2. The larger the $|RMSE_{diff}|$ (absolute value of evaluation performance), the worse the evaluation of the corresponding CV method. In addition, when $RMSE_{diff} < 0$, it indicates the corresponding CV method's evaluation is pessimistic. Conversely, when $RMSE_{diff} > 0$, this CV method's evaluation is optimistic.

$$RMSE_{diff} = RMSE_{actual} - RMSE_{CV} \qquad (2)$$

To obtain $RMSE_{diff}$, we should calculate $RMSE_{actual}$ and $RMSE_{CV}$. For $RMSE_{actual}$, we first train an RF prediction model using all 1000 samples of the specified prediction task, and then apply this RF model to predict all prediction locations (i.e. all grids of the dataset except these 1000 samples) to obtain their predicted values of the target variable. Based on all prediction locations' true and predicted values (of target variable), we can calculate $RMSE_{actual}$ of this specified prediction task. For $RMSE_{CV}$, we implement all CV methods (RDM-CV, BLK-CV, and SP-CV) based on all 1000 samples of the specified prediction task (detailed steps are introduced in Subsection 2.2.), and then, we obtain their $RMSE_{CV}$s. Finally, for each prediction task, we calculate $RMSE_{diff}$ of each CV method according to Equation 2.

### 3.2.3 Step 3: Plot the relationship of dissimilarity and CV methods evaluation performances.

In summary, following steps 1 and 2, we have 1000 prediction tasks for each dataset. Each prediction task corresponds to a unique dissimilarity value and is used to obtain the evaluation performances ($RMSE_{diff}$s) of three CV methods. By plotting the dissimilarity vs $RMSE_{diff}$, we can analyze the relationship between dissimilarity and evaluation performances of CV methods.

Our results are presented in Figure 9. The y-axis represents the value $RMSE_{diff}$, and the x-axis represents the value of dissimilarity. In each scatter plot, there are 3000 points that correspond to $RMSE_{diff}$s of the 1000 predictions linked to each of the three CV methods considered in this research. Points around the zero line (x-axis) correspond to accurate evaluations. Points below and above that line represent pessimistic and optimistic evaluations, respectively.

Figure 9 confirms the results presented by recent studies (de Bruin et al. 2022; Milà et al. 2022; Wadoux et al. 2021): RDM-CV is over-optimistic when sample data and prediction locations are different, while geospatial CV methods tend to be over-pessimistic when samples almost cover the entire prediction area. In Figure 9, it is obvious that RDM-CV points are clearly above the zero line in large dissimilarity values, and it is also worth noting that SP-CV points correspond to pessimistic evaluations in low dissimilarity values.

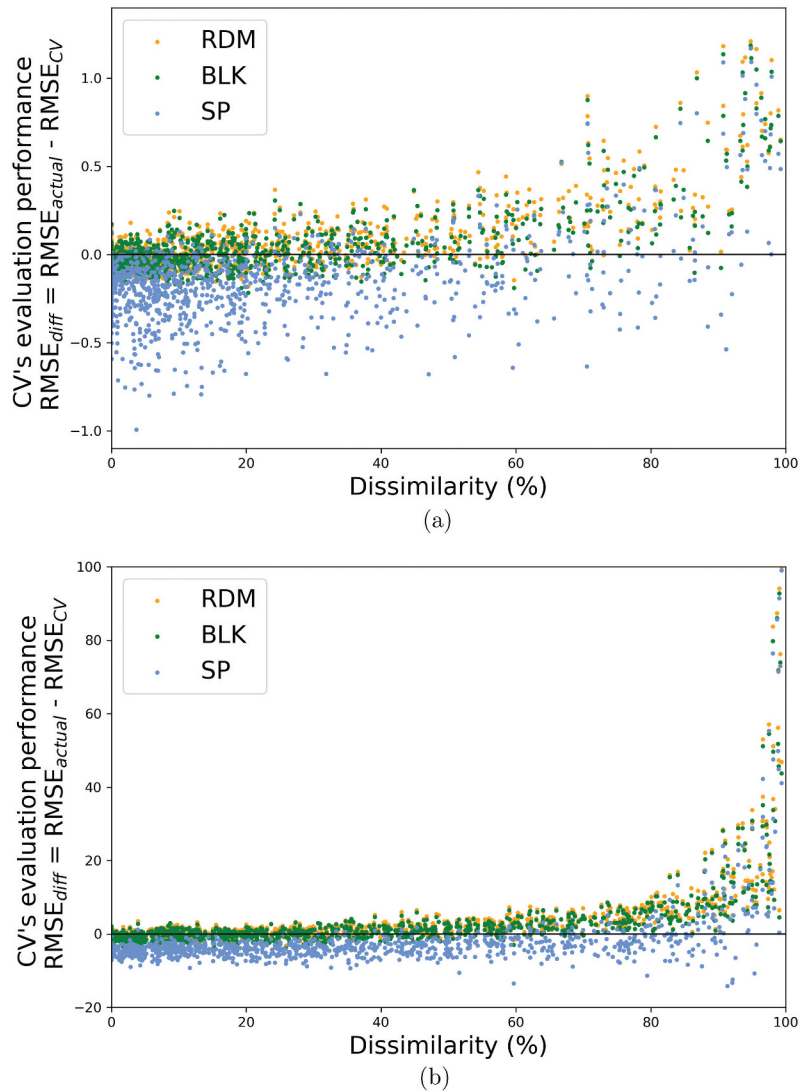

(a)

(b)

**Figure 9.** Final scatter plots of experiments. X-axis: dissimilarity values. Y-axis: CV method evaluation performance ($RMSE_{diff}$) values. (a). Synthetic dataset. (b). Real dataset.

Furthermore, Figure 9 provides new insights into the relationship between dissimilarity and CV evaluation performance ($RMSE_{diff}$). Unlike previous studies, which only analyzed a few dissimilarity degrees, here we explore gradually changing dissimilarities. Firstly, we observe that over-optimistic RDM-CV and over-pessimistic geospatial CV methods could happen simultaneously in the intermediate dissimilarity scenarios. This finding further reinforces the argument put forth by Wadoux et al. (2021), suggesting that neither RDM-CV nor geospatial CV methods are suitable for evaluating geospatial ML predictions, particularly the presence of diverse dissimilarity scenarios. Secondly, the variations in $RMSE_{diff}$ are not uniform across all dissimilarities. As dissimilarity increases, the rate of $RMSE_{diff}$ change also increases. This discovery serves as a significant addition to comprehensively understanding the relationship between dissimilarity and CV evaluation performance.

Because it is hard to read scatter plots with 3000 points, we binned all dissimilarities to the 1% (i.e. we create 100 bins from the original experiments). After that, the corresponding $|RMSE_{diff}|$ of each bin is calculated by averaging the absolute values in the bin. Note that direct averaging $RMSE_{diff}$ values could lead to positive and negative values, cancelling each other out. Hence, we average absolute values to obtain effective statistical results. The results of this operation are depicted in Figures 10a,b, where we see three rough intervals based on the dissimilarity values. The first one is [0%, 50%). In this interval, SP-CV is appreciably worse than the other CV methods and RDM-CV seems to provide almost unbiased evaluations, especially in the first half of this interval. When the dissimilarity is larger than 30%, BLK-CV is slightly better than RDM-CV. Generally speaking, when dissimilarity was low, RDM-CV usually had a more accurate evaluation than geospatial CV methods. For example, when dissimilarity is around from 15% to 25%, the $|RMSE_{diff}|$ points of RDM-CV were closer to the zero line in both Figures 10a,b, indicating RDM-CV was more accurate. This result was consistent with the experimental results of Wadoux et al. (2021) and Milà et al. (2022).

The second interval is [50%, 90%). In this interval, the $|RMSE_{diff}|$ of SP-CV gradually becomes better. However, when dissimilarity is between 50% and 80%, the $|RMSE_{diff}|$s of RDM-CV and geospatial CV methods (BLK-CV and SP-CV) are all less than satisfactory, and it is not clear which method is certainly more accurate. Until dissimilarity surpasses 80% and below 90%, SP-CV becomes notably superior to other CV methods. This suggests that the consideration of feature space in SP-CV plays an important role, especially when there are substantial differences between sample data and prediction locations. For instance, when dssimilarity is from 75% to 85%, We could find that $|RMSE_{diff}|$ points of SP-CV were the closest to the zero line and BLK-CV were usually closer to the zero line than RDM-CV in both Figures 10a,b. It indicates that geospatial CV methods were more accurate than RDM-CV in this condition, which aligns with the conclusion of Wadoux et al. (2021) and Milà et al. (2022) too.

In the third and last intervals (i.e. [90%, 100%], the dissimilarity between sample data and prediction locations is too large and none of the CV methods provides acceptable $|RMSE_{diff}|$s, with them all being over-optimistic.

To gain a deeper understanding of how the evaluation performances of CV methods changes with dissimilarities, the scatter plots of $RMSE_{actual}$s and $RMSE_{CV}$s are put together in Figures 11a,b. In these figures, it is clearly noticeable that the variations in $RMSE_{actual}$s are much greater than the changes of $RMSE_{CV}$ of three CV methods. Consequently, the differences of CV evaluation performances in diverse dissimilarity scenarios are mainly due to the variations of actual prediction errors. RDM-CV and geospatial CV methods are not capable of reflecting changes in dissimilarity, which results in that they cannot consistently provide accurate evaluations in diverse dissimilarity scenarios. Figures 11a,b also show that $RMSE_{CV}$s of SP-CV are consistently higher than that of BLK-CV and RDM-CV and that the $RMSE_{CV}$s of BLK-CV are slightly higher than those of RDM-CV. In other words, geospatial CV methods provide higher $RMSE_{actual}$s reflecting that they indeed have the ability to better simulate the difference between sample data and prediction locations.

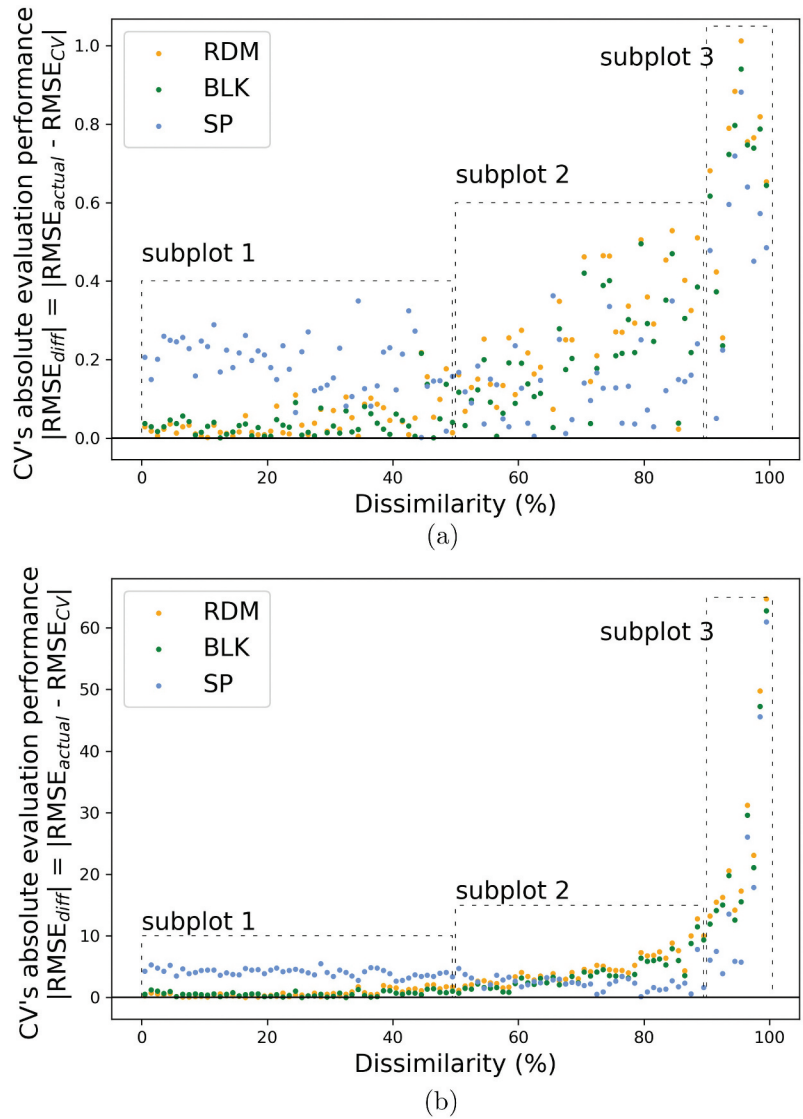

**Figure 10.** Final scatter plots with binned dissimilarities. X-axis: dissimilarity values (with 100 bins). Y-axis: CV method absolute evaluation performance ($|RMSE_{diff}|$) values. (a). Synthetic dataset. (b). Real dataset. Interval of dissimilarity in subplot 1: [0%, 50%), in subplot 2: [50%, 90%), in subplot 3: [90%, 100%].

In addition, the changing pattern of SP-CV $RMSE_{CV}$ in the dissimilarity range [60%, 90%) is different from that of RDM-CV and BLK-CV. In this range, SP-CV shows a relatively stable behavior, while RDM-CV and BLK-CV show rapidly decreasing evaluation results. Another interesting pattern is observed in the dissimilarity range of [90%, 100%] where we see that the $RMSE_{CV}$ of SP-CV rapidly decreases. This is mainly because the spatial coverage of samples in this context is too small, and the configured sample data lack sufficient internal variation. As a result, SP-CV could not completely reflect the dissimilarity in this range.

Figures 10 and 11 also show that experimental results of synthetic and real datasets are not completely identical. The most obvious difference is that the points of Figures 10b and 11b are much less fluctuant than those of Figures 10a and 11a, demonstrating that actual prediction error and CV methods' evaluations are more stable in the real dataset experiment. The fundamental reason for this difference in stability lies in the correlation between the covariates and the target variable. Specifically, in the synthetic dataset, as mentioned in Subsection 3.1 datasets, only three out of seven covariates are correlated to the target variable.

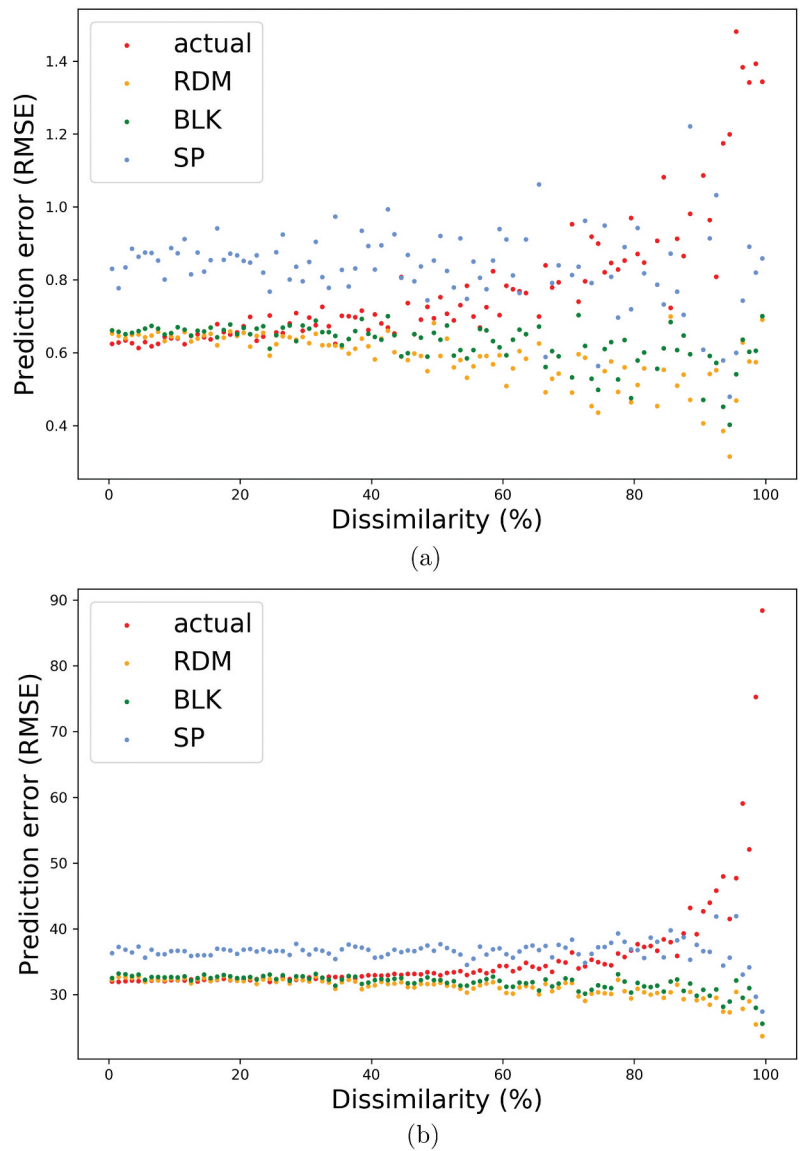

**Figure 11.** Final scatter plots of prediction error with binned dissimilarities. X-axis: dissimilarity values (with 100 bins). Y-axis: Prediction error ($RMSE_{actual}$ and $RMSE_{CV}$) values. (a). Synthetic dataset. (b). Real dataset.

However, in the real dataset, 28 covariates in total are all correlated with the target variable. Therefore, the prediction ability of the ML model of the real dataset is more stronger than the one of the synthetic dataset, of course leading to much more stable results across different prediction tasks. Finally, such stability also appears in the experimental results.

Although synthetic and real datasets' experiments have differences, the relationship between dissimilarity and CV evaluations exhibits similar trends and considerable commonalities. This is why in the above discussions we do not distinguish between the two datasets. These commonalities demonstrate the effectiveness and versatility of the proposed method to quantify dissimilarity in different geospatial ML predictions. They also demonstrate that the impact of dissimilarity on CV methods' performances roughly follows similar patterns.

## 4. Conclusions and future research

With the advancement of geographical ML predictions, researchers have recognized the importance of dissimilarity between sample data and prediction locations and its crucial role in the evaluation of such predictions. However, there is a lack of methods to quantify this dissimilarity, which could also be used to help select a suitable CV evaluation method. Here, we propose dissimilarity quantification by adversarial validation (i.e. DAV) based on the information contained in the feature space.

DAV was tested using a series of prediction tasks with gradually changing dissimilarity degrees and using both synthetic and real datasets. Results showed that DAV could effectively provide corresponding dissimilarities to the geospatial ML predictions. We also compared RDM-CV and two representative geospatial CV methods (BLK-CV and SP-CV) in the experiments to investigate how dissimilarity affects the evaluation performance of CV methods. Our results presented that neither random CV nor geospatial CV methods can consistently provide accurate evaluations across a range of dissimilarity degrees. Therefore, we suggest designing "self-adaptive" CV methods that future work can concentrate on, providing accurate evaluations in a much wider dissimilarity range.

DAV has great potential in broader geoscience applications. For example, DAV can also be used to quantify the dissimilarity between classification and semantic segmentation of remote sensing images. DAV and its quantified dissimilarity can also be used to design sampling strategies, optimize prediction models, and improve CV methods or train-validation-test split. In the future, we also plan to apply DAV in more applications and datasets and employ more ML classifiers in DAV, to further verify the availability and generalizability of DAV.

## Disclosure statement

## Funding

The work was supported by the China Scholarship Council [201804910723].

## ORCID

Yanwen Wang  http://orcid.org/0000-0001-8070-2122
Mahdi Khodadadzadeh  http://orcid.org/0000-0001-7899-738X
Raúl Zurita-Milla  http://orcid.org/0000-0002-1769-6310

## Data and code availability